\pdfoutput=1

\documentclass[11pt]{article}

\usepackage{emnlp2021}

\usepackage{times}
\usepackage{latexsym}

\usepackage[T1]{fontenc}

\usepackage[utf8]{inputenc}

\usepackage{microtype}

\usepackage{times}
\usepackage{graphicx}
\usepackage{multicol}
\usepackage{multirow}
\usepackage{mathrsfs}
\usepackage{amsmath}
\usepackage{amssymb}
\usepackage{subfigure}
\usepackage{latexsym}
\usepackage{booktabs}

\usepackage[linesnumbered,ruled]{algorithm2e}

\SetCommentSty{mycommfont}

\definecolor{layer1}{RGB}{64,116,155}
\definecolor{layer2}{RGB}{112,48,160}

\title{CR-Walker: Tree-Structured Graph Reasoning and Dialog Acts for Conversational Recommendation}

\author{Wenchang Ma$^{1*}$, Ryuichi Takanobu$^{2*}$, Minlie Huang$^{1\dagger}$\\
$^{1}$CoAI Group, DCST, IAI, BNRIST, Tsinghua University, Beijing, China\\
$^{2}$Alibaba Group, Hangzhou, China\\
{\small \tt mwc17@mails.tsinghua.edu.cn, ryuichi.gxly@alibaba-inc.com, aihuang@tsinghua.edu.cn}
}

\begin{document}
\maketitle
\begin{abstract}
Growing interests have been attracted in Conversational Recommender Systems (CRS), which explore user preference through conversational interactions in order to make appropriate recommendation. However, there is still a lack of ability in existing CRS to (1) traverse multiple reasoning paths over background knowledge to introduce relevant items and attributes, and (2) arrange selected entities appropriately under current system intents to control response generation. 
To address these issues, we propose CR-Walker in this paper, a model that performs tree-structured reasoning on a knowledge graph, and generates informative dialog acts to guide language generation. The unique scheme of tree-structured reasoning views the traversed entity at each hop as part of dialog acts to facilitate language generation, which links how entities are selected and expressed.  Automatic and human evaluations show that CR-Walker can arrive at more accurate recommendation, and generate more informative and engaging responses.
\end{abstract}
\renewcommand{\thefootnote}{\fnsymbol{footnote}}
\footnotetext[1]{Equal contribution.}
\footnotetext[2]{Corresponding author.}
\renewcommand{\thefootnote}{\arabic{footnote}}

\section{Introduction}
Many researches have been drawn to combine conversational assistants with recommender agents into one framework due to its significance and value in practical use \cite{sun2018conversational,jannach2020survey}, but creating a conversational recommender system (CRS) that bridges conversation and recommendation still remains challenging. 

\begin{figure}[htb]
    \centering
    \includegraphics[width=1.05\linewidth]{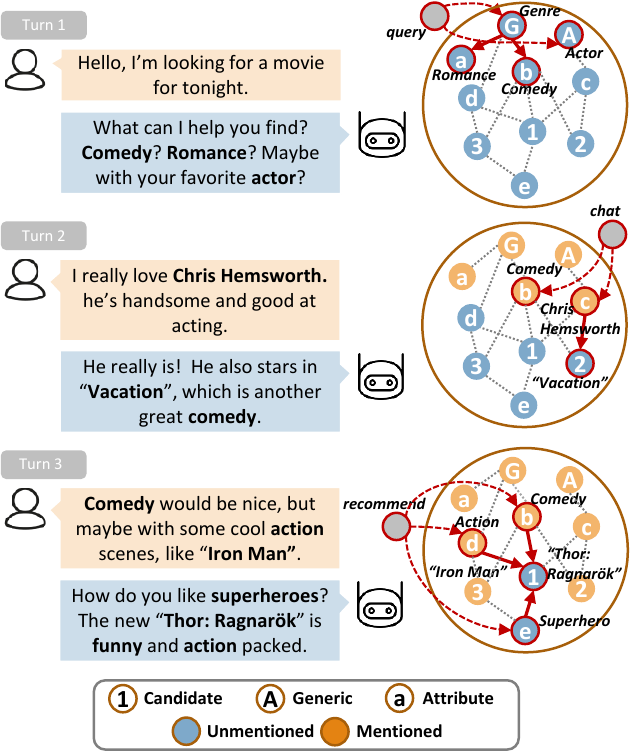}
    \caption{First three turns of an example dialog. The dialog is shown on the left with entities on the KG in bold. The graph on each dialog turn's right demonstrates the reasoning process of CR-Walker, with the reasoning tree marked red. Throughout this paper, candidate items are noted with numbers, and generic classes / attributes with upper-/lower-case letters. The orange/blue color indicates that the entity is mentioned/unmentioned in the previous context.}
    \label{fig:toy} 
\end{figure}

One of the challenges lies in \textbf{reasoning over the background knowledge for accurate recommendation}. Prior studies usually focused on using context and knowledge as an enrichment to recommendation \cite{chen2019towards,zhou2020improving}, but neglected to fully exploit the connection between entities\footnote{An entity can be any node on a knowledge graph throughout this paper, including items and their attributes. The definition is provided in Sec. \ref{sec:concept}.} to infer the system action. In particular, this requires the system to make \textit{multi-path reasoning} on background knowledge, since one entity may have multiple relationships with different entities through multi-hop reasoning. For example in Fig. \ref{fig:toy}, after the user mentioned ``Hemsworth'', the agent chats on ``Vacation'' starring ``Hemsworth'', and further explores the user interests in ``Comedy'' movies. It then recommends ``Thor'' based on several distinct paths of reasoning over user preference (``comedy'' \& ``action'').

Another challenge lies in \textbf{fully utilizing the selected entities in response generation}. Since different dialog actions can be applied in conversational recommendation, selected entities needs to be properly expressed with the guide of \textit{dialog acts}, an abstract representation of dialog semantics and intentions, in order to form natural, informative, and engaging utterances to interact with users. However, most previous works \cite{moon2019opendialkg,lei2020estimation} stopped at inferring entities without modeling response generation. In Fig. \ref{fig:toy} again, the agent first asks the user's preferred genres and actors, and then talks about the star and the movie to engage the user in the conversation, and last recommends a movie based on the user interests. In addition, the agent provides explanations at the third turn to make recommendation more interpretable and persuasive.

To address these issues, we propose \textbf{C}onver-sational \textbf{R}ecommendation \textbf{Walker} (CR-Walker) in this paper. It first selects a system intent to decide whether the system asks for information, chats about something, or makes a recommendation. Then, it performs tree-structured reasoning on a knowledge graph (KG) and dialog context, creating a reasoning tree comprised of relevant entities to be introduced in response. The hierarchical arrangement of entities on the tree preserves the logical selection order under current system intents, which is transformed to dialog acts. 
The linearized representation of dialog acts further guides on generating informative and engaging responses with pre-trained language models.
Results show that CR-Walker outperforms strong CRS on two public datasets in recommendation and generation tasks.

In brief, our contributions are summarised below\footnote{The codes are released at \url{https://github.com/truthless11/CR-Walker}}: 
    (1) CR-Walker conducts tree-structured reasoning on a knowledge graph and dialog context to explore background knowledge and exploit connection between entities for more accurate recommendation;
    (2) CR-Walker transforms the reasoning tree into dialog acts that abstract the semantics and hierarchy of selected entities, and thereby generates more engaging responses for recommendation; 
    (3) We evaluate CR-Walker on two conversational recommendation datasets, achieving outstanding performance in automatic and manual evaluation, from both recommendation and conversation aspects.

\section{Related Work}\label{sec:related}

Conversational Recommender Systems (CRS) learn and model user preference through dialog, which support a richer set of user interactions in recommendation \cite{jannach2020survey}. Previous CRS can be roughly categorized into two types. 

One is recommendation-biased CRS \cite{sun2018conversational,zhang2018towards,zhang2020conversational,zou2020towards} that asks questions about user preference over pre-defined slots or attributes to recommend items. 
As system response can be grouped into some pre-defined intents, it can be implemented with the help of language templates. Under this simplified setting, approaches of this type do not model language generation explicitly \cite{lei2020estimation,lei2020interactive}. Such dialogs can only provide limited actions without revealing why the system makes such recommendation (e.g. by asking on a fixed set of attributes) \cite{christakopoulou2016towards,christakopoulou2018q}, thus leading to unsatisfactory user experience. Recently, \citet{moon2019opendialkg} improves knowledge selection by assuming a single chain of reasoning throughout the conversation. It relies on fine-grained annotations that follow single-path reasoning scheme. However, multiple entities can be selected at each reasoning hop (e.g. recommend several items within one turn, each item with different reasons). Therefore, we propose tree-structured reasoning in this work to enable CRS to select multiple entities through multi-path reasoning for accurate recommendation. \citet{xu2020user} introduces a dynamic user memory graph to address the reasoning of user knowledge in CRS, which is beyond the scope of this paper.

\begin{figure*}[htb]
  \centering
  \includegraphics[width=\textwidth]{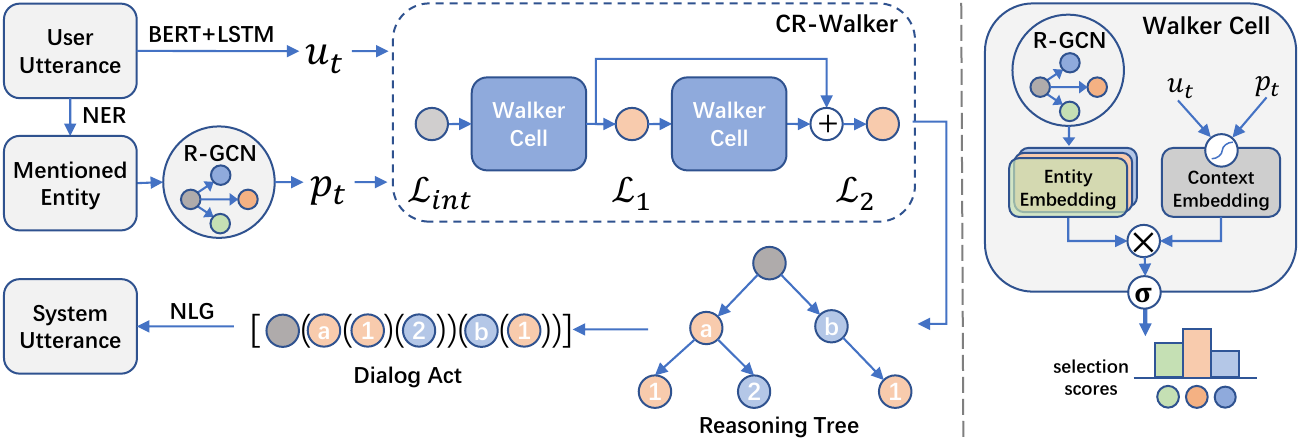}
  \caption{Left: Illustration of CR-Walker's overall architecture. CR-Walker first decides the system intent and then applies walker cells to perform tree-structured reasoning on the knowledge graph in two stages. The transformed dialog acts are used to guide response generation. Right: Detailed structure for a single walker cell. A walker cell calculates the similarity between the entities on a graph and the context embedding that integrates utterance embedding and user portrait. Entity selection is learned by logistic regression to enable multiple selections.}
  \label{fig:crwalker}
\end{figure*}

The other is dialog-biased CRS \cite{li2018towards,kang2019recommendation,liao2019deep,liu2020towards} that makes recommendations using free text, which have much flexibility to influence how the dialog continues. 
As these systems suffer from existing limitations in NLP (e.g. understand preference implicitly from user expression), most methods incorporate external information such as KG and user logs to enhance the dialog semantics \cite{yu2019visual,zhou2020improving} or update the user representation \cite{zhang2019text,chen2019towards}. However, these methods do not capture higher-level strategic behaviors in recommendation to guide the conversation.
To solve this issue, \citet{zhou2020towards} incorporates topic threads to enforce transitions actively towards final recommendation, but it models CRS as an open-ended chit-chat task, which does not fully utilize relations between items and their attributes in response.
In contrast, CRS can be regarded as a variation of task-oriented dialog system that supports its users in achieving recommendation-related goals through multi-turn conversations \cite{tran2020deep}. 
Inspired by the use of dialog acts \cite{traum1999speech}, we choose a set of system dialog acts in CRS to facilitate information filtering and decision making as task-oriented dialog policy \cite{takanobu2019guided,takanobu2020multi} does.

\section{CR-Walker: Conversational Recommendation Walker}
In this section, we start from defining the key concepts of knowledge graph and dialog acts used in CR-Walker. As illustrated in Fig. \ref{fig:crwalker}, CR-Walker works as follows: First of all, dialog history is represented in two views: one is utterance embedding in the content view, 
and the other is user portrait in the user interest view.
Then, CR-Walker makes reasoning on a KG to obtain a reasoning tree, which is treated as a dialog act. Afterwards, the tree-structured dialog act is linearized to a sequence, on which CR-Walker finally generates responses with a conditional language generation module. 

\subsection{Key Concepts}\label{sec:concept}
We construct a knowledge graph $ G = (\mathcal{E}, \mathcal{R}) $ as follows: the entities $\mathcal{E}$ on the graph are divided into three categories, namely \textbf{candidate items}, \textbf{attributes}, and \textbf{generic classes}. There are various relations $\mathcal{R}$ among these entities. Each candidate item is related to a set of attributes, while each attribute is connected to its corresponding generic class. There might also exist relationships between different attributes. Taking movie recommendation as an example, the candidate movie \textit{Titanic} is linked to attributes \textit{Romance}, \textit{Leonardo DiCaprio} and \textit{James Cameron}, and these three attributes are linked to generic classes \textit{Genre}, \textit{Actor} and \textit{Director}, respectively. 

We also define a set of system actions in CRS. We abstract three different system intents to represent actions commonly used in a dialog policy: \textbf{recommendation} that provides item recommendation and persuades the user with supporting evidence, \textbf{query} that asks for information to clarify user needs or explore user preference, and \textbf{chat} that talks on what has been mentioned to drive the dialog naturally and smoothly. Example utterances of three intents are shown in Fig. \ref{fig:toy}. Then, we define a dialog act $A$ as an assembly of a system intent and entities selected by the system, along with their hierarchy relations.

\begin{table}[htb]
    \centering
    \small
    \begin{tabular}{lcc}
    \toprule
        Intent & Hop 1 & Hop 2 \\
    \midrule
        Recommend & mentioned attributes & candidate items \\
        \multicolumn{3}{c}{+ mentioned items' attributes ~~~~~~} \\
        \cmidrule(lr){1-3}
        Query & generic classes & attributes\\
        \cmidrule(lr){1-3}
        Chat & mentioned entities & all entities\\
    \bottomrule
    \end{tabular}
    \caption{Reasoning rules for narrowing down CR-Walker's search space. For each system intent, we only maintain the legal entities at each hop during reasoning.}
    \label{tab:reasoning}
\end{table}

\subsection{Reasoning Process}\label{sec:reasoning}
CR-Walker learns to reason over KG to select relevant and informative entities for accurate recommendation and generating engaging conversations. Considering the large scale of KG and different system actions in CRS, we design several two-hop reasoning rules to help CR-Walker narrow down the search space, thereby making the reasoning process more efficient on large KG.  
As shown in Table \ref{tab:reasoning}, all the reasoning rules are designed in line with the conceptual definition of corresponding intents. 
The reasoning process of CR-Walker starts from one of the three intents. It then tries to explore intermediate entities as a bridge to the final recommendation, and finally reaches the target entities at the second hop.

As explained in Sec. \ref{sec:related}, multiple entities can be selected at each hop in CRS, therefore forming a tree structure on the graph instead of a single path as in previous work \cite{moon2019opendialkg}. The child entities at the second hop are \textit{neighboring} to their parent entities at the first hop on the graph, except when the intent is ``recommend''. We allow all candidate items to be recommended, even if some of them have no connection with other entities on the graph. In addition, we maintain the status of each entity whether the entity is mentioned or not in the context, to facilitate reasoning during interaction.

\subsection{Dialog and Knowledge Representation}
In this subsection, we describe how to represent dialog context, external knowledge and user interests in CR-Walker.

\paragraph{Utterance Embedding}
We formulate the dialog history $D = \{ x_1, y_1, \dots, x_{t-1}, y_{t-1}, x_t \}$, where $x_t$ and $y_t$ is user/system utterance respectively. At each dialog turn $t$, we first use BERT \cite{devlin2019bert} to encode last system utterance $y_{t-1}$ and current user utterance $x_t$ successively. The embedding of ``[CLS]'' token of $x_t$ is applied as the turn's representation, denoted as $\text{BERT}([y_{t-1};x_t])$. Then the utterance embedding $\boldsymbol{u}_t$ is obtained simply through a LSTM over $\text{BERT}([y_{t-1};x_t])$ to capture the sentence-level dependencies. 
Formally,
\begin{align}\label{eq:utterance}
    \boldsymbol{u}_t= \text{LSTM}(\boldsymbol{u}_{t-1},\text{BERT}([y_{t-1};x_t])).
\end{align}
The hidden state of LSTM $\boldsymbol{u}_t \in \mathbb{R}^d$ is taken as the utterance embedding to represent dialog context. 

\paragraph{Entity Embedding}
To introduce external structured knowledge in CR-Walker, we extract KG from DBpedia \cite{auer2007dbpedia} and add generic classes (see Sec. \ref{sec:concept}). We encode the graph using R-GCN \cite{schlichtkrull2018modeling}, by virtue of its capability of modeling neighboring connections more accurately by considering different relations. Formally, for each entity $e \in \mathcal{E}$, the entity embedding $\boldsymbol{h}_e^{(l)} \in \mathbb{R}^{d}$ at each layer $l$ is calculated as:
\begin{align}
    \boldsymbol{h}_e^{(l+1)}=\sigma(\sum_{r\in \mathcal{R}}\sum_{e'\in \mathcal{N}_e^r}{\frac{1}{|\mathcal{N}_e^r|}}\boldsymbol{W}_r^{(l)}\boldsymbol{h}_{e'}^{(l)}+\boldsymbol{W}_0^{(l)}\boldsymbol{h}_e^{(l)}), 
\end{align}
where $\mathcal{N}_e^r$ denotes the set of neighboring entities of $e$ under the relation $r$, and $\boldsymbol{W}_r^{(l)}$,$\boldsymbol{W}_0^{(l)} \in \mathbb{R}^{d \times d}$ are learnable matrices for integrating relation-specific information from neighbors and the current layer's features respectively. 
At the final layer $L$, the embedding $\boldsymbol{h}_e^{(L)}$ is taken as the entity representation, and is denoted as $\boldsymbol{h}_e \in \mathbb{R}^d$ in the following text. 

\paragraph{User Portrait}
We build a user portrait to represent user interests using both dialog and KG here. Given the dialog history, we performed \textit{named entity recognition} (NER) to identify entities mentioned in the previous user utterances $\{x_1,\dots,x_{t-1},x_t\}$ using spaCy, 
then linked them to the entities in the KG with simple fuzzy string matching. 
The status of identified entities is updated as ``mentioned''. We thus obtain all the representation of mentioned entities $\boldsymbol{M}_t \in \mathbb{R}^{d\times|M_t|}$ by looking up entity embedding:
\begin{align*}
    \boldsymbol{M}_t=(\boldsymbol{h}_1,\boldsymbol{h}_2,...,\boldsymbol{h}_{|M_t|}).
\end{align*}
Following \citet{chen2019towards}, we calculate the user portrait $\boldsymbol{p}_t \in \mathbb{R}^d$ via self-attention:
\begin{align} \label{eq:portrait}
    \boldsymbol{\alpha}_t&=\text{softmax}(\boldsymbol{w}_p\cdot \text{tanh}(\boldsymbol{W}_p\boldsymbol{M}_t)), \notag \\ 
    \boldsymbol{p}_t&=\boldsymbol{\alpha}_t*\boldsymbol{M}_t. 
\end{align}

\subsection{Tree-Structured Graph Reasoning}
The reasoning process of CR-Walker initiates from a system intent as the start point on the graph, and expands into multiple paths to get a reasoning tree. 

First of all, we treat intent selection as a simple 3-way classification problem parameterized by $\theta_i$:
\begin{align}\label{eq:intent}
    p_{\theta_i}(y_t^{int}|x_t)&=\text{softmax}(\boldsymbol{W}_{int}^2\text{ReLU}(\boldsymbol{W}_{int}^1\boldsymbol{u}_t)), \notag \\
    \mathcal{L}_{int}&= -\log p_{\theta_i}(y_t^{int}|x_t).
\end{align}
Noting that we only use utterance embedding $\boldsymbol{u}_t$ as input, since we empirically find that introducing $\boldsymbol{p}_t$ does not promote the performance of intent selection.

To expand a system intent into a reasoning tree, we propose the \textit{walker cell}, a neural module shown in Fig. \ref{fig:crwalker}. Particularly, each time a walker cell $C$ performs one-hop reasoning to select entities, to expand the tree from a given intent $i$, or a given entity $e$ at hop $n$=1 or >1 respectively. It first integrates the dialog history representation via a gate mechanism to obtain context embedding $\boldsymbol{c}_t$:
\begin{align}
    \gamma^{(n)}&=
\begin{cases}
    \sigma(\boldsymbol{w}\cdot[\boldsymbol{u}_t;\boldsymbol{p}_t;\boldsymbol{i}_t]), & n\text{=1} \\ 
    \sigma(\boldsymbol{w}\cdot[\boldsymbol{u}_t;\boldsymbol{p}_t;\boldsymbol{i}_t;\boldsymbol{h}_e]), & n\text{>1}
\end{cases}
 \notag \\
    \boldsymbol{c}_t^{(n)}&=\gamma^{(n)}\cdot\boldsymbol{u}_t+(1-\gamma^{(n)})\cdot\boldsymbol{p}_t,
\end{align}
where $\boldsymbol{i}_t \in \mathbb{R}^d$ indicates trainable embedding of the selected intent. The cell then outputs the score of each entity $e'$ using its entity embedding $\boldsymbol{h}_{e'}$:
\begin{align}\label{eq:score}
    \hat{s}_{e'}= \sigma(\boldsymbol{h}_{e'}\cdot \sum_{j=1}^n\boldsymbol{c}^{(j)}_t).
\end{align}
The estimated selection score $\hat{s}_{e'}$ indicates whether $e'$ is selected for tree expansion. By incorporating $\boldsymbol{c}_t^{(j<n)}$, the current reasoning hop $n$ is aware of the previous reasoning hop $j$. We describe this process of applying a single walker cell for entity selection from $e$ (similar for intent $i$) as a function below:
\begin{align}\label{eq:reasoning}
    \text{WALK}(e)=\{e' | \hat{s}_{e'} > \tau, e' \in \mathcal{Z}_e^{(n)} \},
\end{align}
where $\mathcal{Z}_e^{(n)}$ is the set of legal entities to be selected starting from $e$ according to the reasoning rule in Sec. \ref{sec:reasoning}, and $\tau$ is a threshold hyper-parameter. 

In practice, we select at most $m$ entities at hop 1 to control the reasoning tree's width. The reasoning path stops when no entities are selected at a certain hop or reaches hop 2.

\subsection{Conditional Language Generation}
Having selected the entities on the reasoning tree, we generate system response $y_t$ conditioned on the user utterance $x_t$ and tree-structured dialog act $A_t$. We formulate this as a language generation problem. The goal is to build a statistical model parameterized by $\theta_g$ as follows: 
\begin{align}\label{eq:clg}
    p_{\theta_g}(y_t|x_t, A_t) = \prod_{k=1}^K p_{\theta_g}(y_{k,t}|y_{<k,t}, x_t, A_t).
\end{align}


To facilitate response generation using a pretrained language model (PLM), we convert the dialog act into a token sequence. As a dialog act of CR-Walker contains an intent and selected entities, and it is arranged in a tree structure, we can linearize the dialog act into a token sequence in the same way that a parser serializes a tree into a string with \textit{preorder traversal}.  As shown in Fig. \ref{fig:crwalker}, the brackets characterize the hierarchy of the dialog act 
with regard to the logical order of entity selection.

In this paper, we employ GPT-2 \cite{radford2019language} as the backbone for response generation, where the model successively encodes the user utterance $x_t$  and sequence dialog act $A_t$ as input, and then decodes the response $y_t$ in an auto-regressive generation process. 
During inference, Top-$p$ sampling \cite{holtzman2020curious} is used for response decoding.

\subsection{Model Optimization}

At each turn $t$, we train the parameters of walker cells $\theta_w$ at each hop $n$ using standard logistic regression loss:
\begin{align}\label{eq:walk}
    \mathcal{L}_n= & \sum_{e\in \mathcal{E}^{(n-1)}_t}\sum_{e' \in \mathcal{Z}^{(n)}_e} -s_{e'}\log(\hat{s}_{e'}) \notag \\ 
    & -(1-s_{e'})\log(1-\hat{s}_{e'}),
\end{align} 
where $s_{e'} \in \{0,1\}$ is the label indicating the selection of entity ${e'}$, and $\mathcal{E}^{(n-1)}_t$ denotes the extracted entity set at dialog turn $t$ at hop $n$-1\footnote{Specially, $\mathcal{E}^{(0)}_t$ is the singleton of selected intent.}. Training the generation model is performed via maximizing the log-likelihood (MLE) of the conditional probabilities in Eq. \ref{eq:clg} over the user utterance:
\begin{align}\label{eq:generation}
    \mathcal{L}_{gen} = -\sum_{k=1}^{K} \log p_{\theta_g} (y_{k,t}|y_{<k,t},x_t, A_t).
\end{align}
Noting that we use the extracted dialog acts in the corpus during training.

We jointly optimize all trainable parameters mentioned above. The final loss for optimization $\mathcal{L}$ is a weighted sum of all losses\footnote{The outlined algorithm and implementation details of CR-Walker are presented in the appendix.}:
\begin{align}\label{eq:loss}
    \mathcal{L}=\mathcal{L}_{int}+\sum_{n=1}^2\lambda_n\mathcal{L}_n+\mathcal{L}_{gen}.    
\end{align}

\section{Experimental Setting}
\subsection{Data}

We use two public conversational recommendation datasets to verify the effectiveness of CR-Walker. (1) ReDial \cite{li2018towards} is collected by crowd-sourcing workers from Amazon Mechanical Turk (AMT). Two paired workers are assigned with a role of either recommender or seeker. At least 4 different movies are mentioned in each conversation. Each movie mentioned in the dialog is annotated explicitly. (2) GoRecDial \cite{kang2019recommendation} is collected in a similar way using ParlAI. In each dialog, each worker is given a set of 5 movies with corresponding descriptions. The seeker's set represents his or her watching history, and the recommender's set represents candidate movies to choose from. The recommender should recommend the \textbf{correct} movie among the candidates to the seeker. We then construct the KG and perform entity linking separately for GoRecDial and Redial.\footnote{The KG construction details and dataset statistics are shown in the appendix.} 

\subsection{Baselines}
We have compared CR-Walker with several strong approaches in Redial:
    (1) ReDial \cite{li2018towards}: a benchmark model of ReDial that applies an autoencoder recommender, a RNN-based NLG and a sentiment prediction module.
    (2) DCR \cite{liao2019deep}: \textit{Deep Conversational Recommender} uses a pointer network to incorporate global topic control and GCN-based recommendation in response generation.
    (3) KBRD \cite{chen2019towards}: \textit{Knowledge-Based Recommender Dialog} enriches user representation with a KG to give responses and recommendation following the user interests.
    (4) KGSF \cite{zhou2020improving}: \textit{KG-Based Semantic Fusion} incorporates both word-oriented and entity-oriented KGs to enhance the data representations in CRS.

We also adopt several conversation recommendation methods as the baselines in GoRecDial:
    (1) BERT \cite{devlin2019bert}: A BERT fine-tuned on GoRecDial, which encodes dialog contexts and movie descriptions. BERT features are used for response retrieval and movie recommendation.
    (2) R-GCN+GPT: A joint model combining a R-GCN \cite{schlichtkrull2018modeling} for movie recommendation with a Transformer-based language model \cite{vaswani2017attention} for response generation. The movies are scored using similar structures within our walker cell by calculating the dot-product between encoder hidden states and R-GCN embeddings.
    (3) GoRecDial \cite{kang2019recommendation}: a benchmark model of GoRecDial, which is trained via multi-task supervised learning and bot-play learning by formulating the recommendation task as a task-oriented game.

\section{Results}

\begin{table}[!th]
    \centering
    \small
    \begin{tabular}{rcccc}
    \toprule
        \multirow{2}{*}[-0.04in]{Model} & \multicolumn{4}{c}{Recommendation} \\
    \cmidrule(lr){2-5} 
        & R@1 & R@10 & R@50 & Cov. \\
    \midrule
        ReDial & 2.3 & 12.9 & 28.7 & 5.8\\
        DCR & 2.7 & 14.8 & 30.6 & 1.5\\
        KBRD & 3.0 & 16.3 & 33.8 & 11.2 \\
        KGSF & 3.9 & 18.3 & \textbf{37.8} & 12.2\\
    \midrule
        CR-Walker & 3.7 & 17.6 & 37.1 & 14.7 \\
        -depth=1  & 2.9 & 14.9 & 34.3 & 11.4\\
        +ConceptNet & \textbf{4.0} & \textbf{18.7} & 37.6 & \textbf{15.2}\\
    \bottomrule
    \end{tabular}
    \caption{Automatic evaluation of recommendation tasks on \textit{ReDial}.}
    \label{tab:rec_redial}
\end{table}

\begin{table}[!th]
    \centering
    \small
    \begin{tabular}{@{}r@{~~~~}c@{~~~~}c@{~~~~}c@{~~~~}c@{~~~~}c@{~~~~}c@{}}
    \toprule
        \multirow{2}{*}[-0.04in]{Model} & \multicolumn{3}{c}{Generation} & \multicolumn{3}{c}{Knowledge} \\
    \cmidrule(lr){2-4} \cmidrule(lr){5-7}
        & BLEU & Dist-2 & Dist-3 & Prec. & Rec. & F1 \\
    \midrule
        ReDial & 21.9 & 3.72 & 5.24 & 28.3 & 23.0 & 24.0 \\
        DCR & 21.9 & 1.91 & 3.12 & 48.1 & 37.7 & 40.8 \\
        KBRD & 22.8 & 4.92 & 9.21 & 42.1 & 33.3 & 35.9 \\
        KGSF & 18.6 & 4.00 & 5.34 & 37.7 & 32.0 & 33.2 \\
    \midrule
        CR-Walker & 26.6 & \textbf{21.2} & \textbf{48.1} & 46.3 & \textbf{60.3} & \textbf{47.7} \\
        -depth=1  & \textbf{28.0} & 19.2 & 40.8 & \textbf{50.0} & 47.7 & 45.1 \\
    \bottomrule
    \end{tabular}
    \caption{Automatic evaluation of generation tasks on \textit{ReDial}.}
    \label{tab:gen_redial}
\end{table}

\begin{table}[!th]
    \centering
    \small
    \begin{tabular}{rcccccc}
    \toprule
         \multirow{2}{*}[-0.04in]{Model} & \multicolumn{4}{c}{Recommendation} & Gen. \\
    \cmidrule(lr){2-5} \cmidrule(lr){6-6}
         & T@1 & T@3 & C@1 & C@3 & BLEU \\
    \midrule
        BERT & 25.5 & 66.3 & 26.4 & 68.3 & 23.9 \\
        R-GCN+GPT & 75.6 & 95.7 & 76.1 & 96.3 & 25.2\\
        GoRecDial & 77.8 & 97.1 & 78.2 & 97.7 & 27.4\\
    \midrule
        CR-Walker & \textbf{81.1} & \textbf{97.5} & \textbf{81.3} & \textbf{98.1} & \textbf{29.6}\\
    \bottomrule
    \end{tabular}
    \caption{Automatic evaluation on \textit{GoRecDial}, where users chat in the context of their watching history.}
    \label{tab:auto_gorecdial}
\end{table}

\begin{figure*}[tb]
    \centering
    \includegraphics[width=\linewidth]{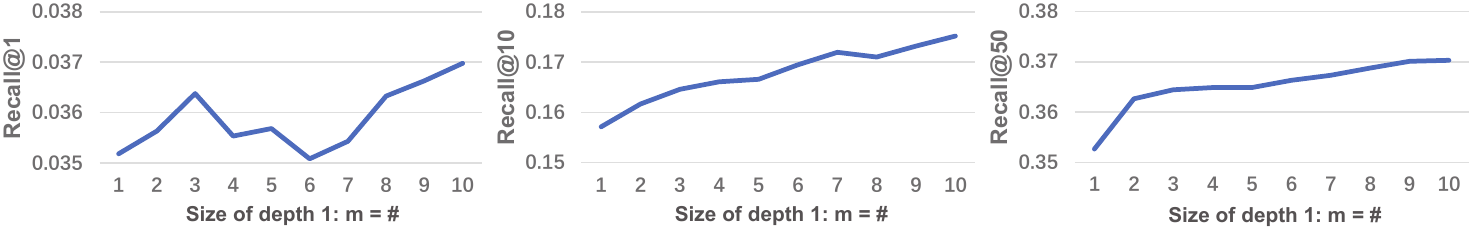}
    \caption{CR-Walker's recommendation performance with regard to the number of selected nodes at the first hop during reasoning. Most metrics improve as more supporting entities are allowed to be selected.}
    \label{fig:size}
\end{figure*}

\subsection{Automatic Evaluation}

The results on Redial and GoRecDial are shown in Table \ref{tab:rec_redial}, \ref{tab:gen_redial} and \ref{tab:auto_gorecdial}. As can be seen,  CR-Walker outperforms most baselines in both item recommendation and response generation. 

\paragraph{Item Recommendation} 
We evaluate CR-Walker on item recommendation quality in different settings using metrics proposed in the original datasets: In Redial, we adopt \textit{Recall@k} for evaluation since there are multiple movies recommended in a dialog. In GoRecDial, since the ground-truth movie to recommend is annotated in each dialog, we evaluate the \textit{hit rate} among top-k recommendation at each turn (\textit{T@k}), and also the hit rate \textit{only at the end of} each dialog (\textit{C@k}) to observe the usefulness of conversation further. On Redial, we also use \textit{Coverage} to evaluate recommendation diversity, which is calculated by the proportion of candidate items recommended on test set.

In Table \ref{tab:rec_redial}, we can find that CR-Walker \textbf{outperformed all baselines using a single KG on recommendation quality}, including ReDial, DCR and KBRD.  This indicates use of multi-path reasoning can more effectively utilize background knowledge. KGSF uses an additional KG from ConceptNet \cite{speer2017conceptnet} compared with ours, and performs slightly better on \textit{Recall}. However, CR-Walker can obtain a performance gain as well by incorporating ConceptNet as additional feature (+ConceptNet in Table \ref{tab:rec_redial}), and even outperforms KGSF on \textit{Recall@1} and \textit{Recall@10}, but this is not the focus of this paper. 
Regarding recommendation diversity, CR-Walker outperformed all baselines including KGSF. The tree structured reasoning enables multiple items to be recommended at the second hop, each with its certain attributes related to earlier conversation. This results in a higher coverage of candidate items compared with 1-hop reasoning that directly arrives at recommendation.

In Table \ref{tab:auto_gorecdial}, we can find that CR-Walker obtains the \textbf{best performance on all recommendation metrics if the user has a clearer preference}. Surprisingly, we also find that T@1 is close to C@1 in CR-Walker in GoRecDial. This is because entity embedding provides overly strong information to distinguish the correct movie from only five candidates so that it can offer good recommendations easily, even without user utterances.

\paragraph{Response Generation}
We apply \textit{BLEU} and \textit{Distinct-n} \cite{li2016diversity} to measure the generated response on word-level matches and diversity. Noting that different from \citet{chen2019towards} that calculate sentence-level Distinct, we use corpus-level Distinct to give a more comprehensive assessment. 
Following \citet{wu2019proactive}, we also adopt \textit{knowledge F1-score} to measure knowledge exploitation. Unlike metrics in item recommendation, the knowledge score is calculated by corresponding generic classes rather than the exact match. For example, it only evaluates whether the system mentioned the \textit{genre} to promote movie recommendation but does not care about the exact genre.

Results show that CR-Walker outperforms all baselines on corpus-level \textbf{language diversity by a large margin} (dist-2,3 in Table \ref{tab:gen_redial}). 
Noticeably, while CR-Walker achieves the highest BLEU in GoRecDial, BLEU in ReDial drops a little when incorporating tree-structured reasoning into the response generation process (26.6 vs. 28.0). This is because CR-Walker sometimes infers different reasoning trees, and afterwards generates sentences that differ from the references but may also be reasonable. We resort to human evaluation (Sec. \ref{sec:human}) to further evaluate the language quality.

In addition, CR-Walker obtains the best knowledge recall and F1 scores. 
This indicates that CR-Walker \textbf{reasonably utilizes informative entities} during conversational recommendation. A slightly lower precision in CR-Walker is also because it produces different reasoning trees. 


\subsection{Ablation Study}

To understand CR-Walker's superiority against other baselines, we further examine the influence of tree-structured reasoning on the recommendation performance. We first study the effect of \textbf{tree depth}. When we simplify the reasoning process by removing the first hop reasoning and force the model to directly predict the entities at the second hop (-depth=1 in Table \ref{tab:rec_redial}), there is an apparent decline in all Recall@k. R-GCN+GPT shares a similar framework with CR-Walker-depth=1 which directly recommends items using R-GCN, and CR-Walker outperforms it much on item recommendation. These results demonstrate that \textbf{two-hop graph reasoning} better exploits the connection between entities by exploring intermediate entities, and it is crucial for successful recommendation. 

We then study the effect of \textbf{tree width}. We control the width of reasoning paths by setting the maximum number of entities $m$ allowed to be selected at the first hop, and observe the performance in Recall@k, as shown in Fig. \ref{fig:size}. Overall, the performance of CR-Walker increases as $m$ goes up. Though there is a slight decrease in Recall@1 when the width grows to around 6, the performance gains in the end. 
This can be interpreted as \textbf{multi-path reasoning} is superior to single-path reasoning by providing the model with multiple guidance to arrive at the final recommendation.

\subsection{Human Evaluation}\label{sec:human}

In addition to automatic evaluation, we conduct point-wise human evaluation. 300 posts are randomly sampled from the test set. For each response generated by each model, we ask 3 worker from AMT to give their ratings according to each metric with a 3-point scale (3/2/1 for good/fair/bad respectively). The average score of each metric is reported. Among the metrics, \textit{fluency} and \textit{coherence} focus on the response generation quality, \textit{informativeness} and \textit{effectiveness} evaluate whether the conversation is well-grounded in a recommendation scenario. In particular, \textit{informativeness} evaluates whether the system introduces rich movie knowledge, and \textit{effectiveness} evaluates whether the system engages users towards finding a movie of interest successfully.

We present human evaluation results on ReDial in Table \ref{tab:human_redial}. We adopt GPT-2 as an additional baseline fine-tuned on the training set and generates response directly. We find that it serves as a solid baseline due to the success of PLMs in language generation and incorporating knowledge implicitly. Although GPT-2 cannot make actual recommendation since it does not ``select'' a movie, it outperforms all the previous baselines even on \textit{informativeness} and \textit{effectiveness}. This implies that finding the appropriate recommendation is insufficient to satisfy users under the conversational recommendation setting, but the quality of natural language may also determine how well recommendations will be accepted. 
CR-Walker, equipping the PLM with external knowledge and reasoning ability, further boosts GPT-2's performance by providing interpretable recommendation through utterance. Among all the metrics, CR-Walker improves informativeness and effectiveness more significantly. We observe that CR-Walker can generate utterance with more detailed attribute information to support recommendation compared to GPT-2 alone. This demonstrates that CR-Walker succeeds in generating engaging responses with tree-structured dialog acts beyond PLMs.

\begin{table}[tb]
    \centering
    \small
    \begin{tabular}{rcccc}
    \toprule
        Model & Fluency & Coherence & Inform. & Effect. \\
    \midrule
        ReDial & 2.31 & 1.96 & 1.69 & 1.74\\
        DCR & 2.12 & 1.84 & 1.68 & 1.63\\
        KBRD & 2.45 & 2.14 & 1.95 & 1.89\\
        KGSF & 2.17 & 1.96 & 1.98 & 1.93\\
        GPT-2 & 2.47 & 2.24 & 2.05 & 1.98\\
        \textit{Human} & 2.52 & 2.34 & 2.18 & 2.10\\ 
        CR-Walker & \textbf{2.60*} & \textbf{2.41*} & \textbf{2.33*} & \textbf{2.22*}\\
    \bottomrule
    \end{tabular}
    \caption{Human evaluation on \textit{ReDial}. 
    \textit{Human} responses come from the ground truth responses provided in the dataset. Numbers marked with * indicate that the improvement is statistically significant (t-test with p-value $<$ 0.05).}
    \label{tab:human_redial}
\end{table}

\begin{table}[tb]
    \centering
    \small
    \begin{tabular}{rcccc}
    \toprule
        Model & Fluency & Coherence & Inform. & Effect. \\
    \midrule
        Human(+) & 2.54 & 2.38 & 2.26 & 2.15\\
        CR-Walker(+) & 2.61 & 2.41 & 2.33 & 2.23\\
    \midrule
        Human(-) & 2.46 & 2.23 & 1.99 & 1.98\\
        CR-Walker(-) & 2.57 & 2.41 & 2.33 & 2.18\\
    \bottomrule
    \end{tabular}
    \caption{Human vs. CR-Walker. (+) and (-) indicate the subset of responses where two competitors share the same intent and pose different intent respectively.}
    \label{tab:human_eval}
\end{table}

We further study why CR-Walker can outperform human responses. 
In terms of the system action, the intent accuracy of CR-Walker reaches only 67.8\%, but we find that a different intent from the human's choice sometimes results in better informativeness and effectiveness.
We calculate the score separately for humans and CR-Walker based on whether the intent selection is the same or different in Table \ref{tab:human_eval}. For identical intents, CR-Walker's improvements on four metrics are all marginal, as the improvement only comes from providing more information at the first hop reasoning. For different intents, however, the human performance drops remarkably, while our performance remains consistent. We observe several samples and find that the human usually performs perfunctory chit-chat like ``haha'' or ``lol'' in these cases. By contrast, CR-Walker replies with a relevant query or appropriate recommendation\footnote{The case study is provided in the appendix.}. This implies that the score advantage may come from the explicit reasoning on system actions that CR-Walker learns.

\subsection{Recommendations in Dialog Flow}

We also analyze the flow of recommended items throughout conversation among various interaction cases, where we roughly categorize the flow into two patterns. In one pattern, the seeker chats around a fixed topic of interest and ask for similar recommendations. This pattern is common on Redial, and CR-Walker efficiently handles it by making appropriate recommendation through tree structure reasoning. However, in a less common case where user suddenly change to a new topic, earlier recommendations would have little effect on the latter ones. In these cases, reasoning on previous items may result in inappropriate recommendations. In practice, we weigh the two patterns by setting the maximum length of dialog history $l_{max}$, where we only used last $l_{max}$ utterances in $D$ to compute utterance embedding and user portrait. When we set $l_{max}=3$, we empirically find CR-Walker can handle most topic changes while still providing appropriate recommendation during interaction.

\section{Conclusion and Future Work}
We have presented CR-Walker, a conversational recommender system that applies tree-structured reasoning and dialog acts. By leveraging intermediate entities on the reasoning tree as additional guidance, CR-Walker better exploits the connection between entities, which leads to more accurate recommendation and informative response generation. Automatic and human evaluations demonstrate CR-Walker's effectiveness in both conversation and recommendation. 
It is worth noting that the dialog acts used in CR-Walker are automatically obtained by entity linking to a KG with simple heuristics. Therefore, our work can be easily applied to different conversational recommendation scenarios.

There are still some topics to be explored based on CR-Walker. It would be promising to equip CR-Walker with a language understanding module to capture users' negative feedback and propose other reasoning rules to handle such situations. An efficient way to learn reasoning more than two hops can be investigated in the future as well.

\section*{Acknowledgement}
This work was supported by the National Science Foundation for Distinguished Young Scholars (with No. 62125604) and the NSFC projects (Key project with No. 61936010 and regular project with No. 61876096). This work was also supported by the Guoqiang Institute of Tsinghua University, with Grant No. 2019GQG1 and 2020GQG0005.

\bibliography{reference}
\bibliographystyle{acl_natbib}

\newpage
\appendix

\section{Notation}
Notations used in this paper are summarized in Table \ref{tab:notation}.
\begin{table}[htb]
    \centering
    \small
    \begin{tabular}{ll}
    \toprule
        Notation & Description \\
    \midrule
        $e, \mathcal{E}$ & Entity, entity set\\
        $r, \mathcal{R}$ & Relation, relation set\\
        $t$ & Dialog turn\\
        $\mathcal{N}_e^r$ & Neighbors of $e$ in relation $r$\\
        $\mathcal{Z}_e^{(n)}$ & Neighbors of $e$ at $n$-th hop reasoning\\
        $\boldsymbol{h}_e$ & Entity embedding of $e$\\
        $\boldsymbol{u}_t$ & Utterance embedding\\
        $\boldsymbol{p}_t$ & User portrait\\
        $\mathcal{E}_t^{(n)}$ & Entity set at $n$-th hop reasoning\\
        $A$ & Dialog act\\
    \bottomrule
    \end{tabular}
    \caption{Notations used in the CR-Walker.}
    \label{tab:notation}
\end{table}

\section{Pseudocode}
The entire reasoning and training process of CR-Walker is described in Algorithm \ref{algorithm}.
\begin{algorithm}[!tb]
\DontPrintSemicolon
\KwIn{Knowledge graph $G$, training data $\mathcal{D}$}
Initialize the parameters of intent classifier $\theta_i$, walker cell $\theta_w$ and generation model $\theta_g$ \;
\For{$j = 1$ \KwTo $|\mathcal{D}|$}{
    Set all the entities on $G$ ``unmentioned'' \;
\For{$t = 1$ \KwTo $T_j$}{
    \tcp{Intent selection}
    Obtain utterance embedding $u_t$ w/ Eq. \ref{eq:utterance} \;
    Calculate $\mathcal{L}_{int}$ based on Eq. \ref{eq:intent} \;
    \tcp{Graph reasoning}
    Obtain user portrait $p_t$ w/ Eq. \ref{eq:portrait} \;
    Update the entities mentioned in user utterances on $G$ as ``mentioned'' \;
    Set $\mathcal{E}_t$ as singleton of the selected intent \;
\For{Hop $n = 1$ \KwTo $2$}{
    Select entities using all elements in $\mathcal{E}_t$ w/ Eq. \ref{eq:reasoning}\;
    Calculate $\mathcal{L}_i$ at current hop w/ Eq. \ref{eq:walk} \;
    Update the selected entities on $G$ as ``mentioned''\;
    Set $\mathcal{E}_t$ as all selected entities at current hop \;
}
    \tcp{Response generation}
    Transform the reasoning tree into the dialog act $A$ \;
    Calculate $\mathcal{L}_{gen}$ w/ Eq. \ref{eq:generation} \;
    Perform gradient descent on $\mathcal{L}$ w/ Eq. \ref{eq:loss} \;
}
}
\caption{Conversational Recommendation Walker}
\label{algorithm}
\end{algorithm}

\section{Implementation Details}

\begin{table}[htb]
    \centering
    \small
    \begin{tabular}{r@{~~}c@{~~}c@{~~}c@{~~}c@{~~}c}
    \toprule
        Dataset & Dialogs & Utterances & Items & Entities & Relations \\
    \midrule
        GoRecDial & 9K & 171K & 3.8K & 19.3K & 227K \\
        ReDial & 10K & 182K & 6.9K & 30.5K & 393K \\
    \bottomrule
    \end{tabular}
    \caption{Dataset statistics.}
    \label{tab:dataset}
\end{table}

\begin{table}[t]
    \centering
    \small
    \begin{tabular}{cccc}
    \toprule
        \textbf{Dataset} & &\textbf{GoRecdial} & \textbf{Redial}\\
    \midrule
        \multirow{7}{*}{\textbf{Entities}} & Movies & 3,782 & 6,924\\
        &Persons & 7,936 & 12,803\\
        &Subjects & 7,558 & 10,707\\
        &Genres & 18 & 18\\
        &Times & 7 & 12\\
        &Generals & 7 & 7\\
        &total & 19,308 & 30,471\\
    \midrule
        \multirow{6}{*}{\textbf{Relations}} &actor\_of & 16,472 & 27,639\\
            &director\_of & 3,634  & 6,063\\
            &genre\_of & 8,386 & 18,259\\
            &subject\_of  & 61,940 & 107,356\\
            &time\_of & 3,782 & 6,309\\
            &is\_a & 19,478 & 30,715\\
            &total & 227,384 & 392,682 \\
    \bottomrule
    \end{tabular}
    \caption{Knowledge graph statistics of \textbf{GoRecdial} and \textbf{Redial}. The total relations are twice the sum of 6 types of relationships listed on the table due to the addition of corresponding reverse relationships.}
    \label{tab:data}
\end{table}

In experiments, we train the model on a single Tesla-V100 GPU with a learning rate of 1e-3, batch size of 36, and max epoch of 60. \textit{Adam} is used as the optimization algorithm, with a weight decay of 1e-2. We set the max number of selection at the first hop $m=5$ during training, and used negative sampling for the candidate items (second hop when system intent is recommend). The ratio between negative and positive samples is set to 5. The dimension of entity embedding $d$ is set to 128. The layer size of R-GCN $L$ is set to 1. BERT-base and GPT-2-medium are applied from \citet{wolf2020transformers} and the parameters of the BERT encoder are frozen during the training process. The weights of graph walker loss at each hop are $\lambda_1=1, \lambda_2 =0.1$ for GoRecDial and $\lambda_1=1, \lambda_2 =1$ for Redial, respectively. During inference, we apply $\tau=0.5$ as the entity selection threshold and $p=0.9$ for the response decoding strategy. Bag of words (BOW) of the movie description are encoded using a fully connected layer as additional features in GoRecDial.

During KG construction, the generic classes we introduce are the \textit{director, actor, time, genre, subject} related to each movie. All entities are directly extracted from DBpedia, except for genres, which are taken from MovieLens. There are 12 types of relationships between entities, namely \textit{actor of} / \textit{director of }/ \textit{genre of} / \textit{subject of} / \textit{time of} / \textit{is a} and their corresponding reverse relationships.  The last relationship \textit{is a} and its reverse appears between attributes and generic classes, while other relationships appear between attributes and candidate items. We then perform entity linking in each sentence, and use regular expressions to determine its intent and entities on the reasoning tree. Noting that no annotation is available in two datasets, and we label them in an automatic way. Dataset statistics are shown in table \ref{tab:dataset}, while detailed entity and relation statistics for KG are listed in table \ref{tab:data}. 



\begin{table}[tb]
    \small
    \begin{tabular}{rp{6.7cm}}
    \toprule
    U: &Hi there, good evening to you!\\
    D: &[ Query ( \textcolor{layer1}{Genre} ) ] \\
    R: &Good evening to you also! What \textcolor{layer1}{kind of films} do you like?\\
    U: &I'm looking for some \textbf{action} flicks like \textbf{Lethal Weapon} to watch with my friends tomorrow. Any ideas?\\
    D: &[ Recommend ( \textcolor{layer1}{Action} ( \textcolor{layer2}{The Terminator} ) ) ( \textcolor{layer1}{1980s} ( \textcolor{layer2}{The Terminator}  ) ) ] \\
    R: &Okay one sec. what about \textcolor{layer2}{The Terminator}. It's \textcolor{layer1}{old} but really great \textcolor{layer1}{action} flick.\\
    U: &\textbf{Terminator}'s great! \textbf{Arnold Schwarzenegger} is a beast in it. We have watched it several times though.\\
    D: &[ Chat ( \textcolor{layer1}{Action} ( \textcolor{layer2}{Die Hard} ) ) ] \\
    R: &Indeed he is, all that muscle. I would also recommend \textcolor{layer2}{Die hard} then. It's a really good mixed \textcolor{layer1}{action} movie.\\
    U: &Great. Can you also give me some \textbf{scary} ones like \textbf{Halloween} or \textbf{Annabelle}?\\
    D: &[ Recommend ( \textcolor{layer1}{Horrow} ( \textcolor{layer2}{Shining} ) ( \textcolor{layer2}{It} ) ) ]\\
    R: &Oh yes! Have you seen \textcolor{layer2}{The Shining} or \textcolor{layer2}{It}?\\
    U: &Yeah... those are really good. We watched \textbf{The Shining} one time and had trouble sleeping...\\
    D: &[ Recommend ( \textcolor{layer1}{Horrow} ( \textcolor{layer2}{Split} ) ( \textcolor{layer2}{Saw} ) ) ] \\
    R: &\textcolor{layer2}{Split} and \textcolor{layer2}{Saw} are even more on the \textcolor{layer1}{scary} side!\\
    U: &Great, I'll check these too. Thanks a lot for the help!\\
    D: &[ Chat ]\\
    R: &No problem. Have a great day!\\
    \bottomrule
    \end{tabular}
    \caption{A sampled case between CR-Walker (R) and a real user (U). The sequential dialog acts (D) at each dialog turn derived from the reasoning tree are presented as well. Entities mentioned in user's turn are marked in \textbf{bold}, while entities in system's turn are marked in colors based on the number of reasoning hop (\textcolor{layer1}{blue}-hop 1, \textcolor{layer2}{purple}-hop 2)}
    \label{tab:case}
\end{table}

\section{Case Study}
We finally present an interactive case here to demonstrate our model's capability during interactive dialog and our model's explainable nature. The sequential dialog acts corresponding to the reasoning tree generated by CR-Walker is presented in Table \ref{tab:case} along with the dialog. We mark all the mentioned entities either in bold (user turn) or in colors (system turn) according to the reasoning hop. The dialog starts with greetings between the user and CR-Walker, followed by CR-Walker proactively seeking user preference by asking which kind of movie he or she likes. The following few turns focus on recommending action movies, and CR-Walker provides an appropriate description of the recommended movies and some comments on \textit{Arnold Schwarzenegger}'s muscles. The topic then switches to horror movies after the user explicitly requires scary ones, with the system recommending four appropriate movies within two turns. The dialog finally ends with the user expressing gratitude and CR-Walker expressing goodwill. Overall, at the utterance level, the whole dialog contains appropriate amounts of information and various dialog acts from the model, enabling the conversation to appear coherent and fluent.

The intermediate dialog acts that CR-walker generates help us to better control and understand the generated utterance. On one hand, the entity on the reasoning tree provides additional insight into the model's particular statement. Generated sentences may contain the entity name directly, but may also contain paraphrase of entities, as in cases of \textit{Genre}, \textit{1980s} and \textit{Horror} mapping to \textit{kind of films}, \textit{old} and \textit{scary} respectively. The model also learns to omit some of the entities on the reasoning path based on the dialog context, such as entity \textit{Horror} when the system recommended \textit{Shining} and \textit{It}. Such non-trivial paraphrasing would be hard to interpret in the absence of the reasoning tree. On the other hand, the reasoning tree's structure even gives a hint to the approach our model takes when it mentions an entity. An interesting case happens in the third turn of the dialog when CR-Walker recommends \textit{Die Hard}. The predicted dialog intent appears to be ``chit-chat'', and \textit{Die Hard} is selected at hop 2 in the reasoning process during inference. As a result, the system talks about the attributes of \textit{Die Hard} (use of \textit{Action}) instead of directly recommending it, and the tone taken by the model is more casual and relevant to the previous context (use of \textit{then} and comment of \textit{all that muscle}). Together, the above advantages add to our model's explainability, giving our model the edge to be interpreted beyond words. 

\end{document}